\documentclass{article}

\usepackage[final]{neurips_2022}
\usepackage{array}
\usepackage{multirow}
\usepackage{booktabs}
\usepackage{caption}

\usepackage[utf8]{inputenc} 
\usepackage[T1]{fontenc}    
\usepackage{hyperref}       
\hypersetup{
    colorlinks=true,
    linkcolor=blue,    
    citecolor=blue,     
    urlcolor=blue      
}
\usepackage{url}            
\usepackage{booktabs}       
\usepackage{amsfonts}       
\usepackage{nicefrac}       
\usepackage{microtype}      
\usepackage{xcolor}         
\usepackage{graphicx}
\usepackage{multirow}
\usepackage{pifont}
\usepackage{colortbl}
\usepackage{subcaption}
\usepackage{lipsum} 
\usepackage{natbib}
\setcitestyle{numbers,square}
\usepackage{url}            
\usepackage{booktabs}       
\usepackage{amsfonts}       
\usepackage{nicefrac}       
\usepackage{microtype}      
\usepackage{xcolor}         
\usepackage{graphicx}
\usepackage{multirow}
\usepackage{pifont}
\usepackage{colortbl}
\usepackage{subcaption}
\usepackage{lipsum} 
\usepackage{natbib}
\usepackage{titlesec}
\usepackage{longtable}
\usepackage{color}
\usepackage{pifont}
\usepackage{longtable}
\usepackage{multirow}  
\usepackage{array}     
\usepackage{geometry}
\setcitestyle{numbers,square}
\bibpunct{[}{]}{,}{n}{,}{,}

\usepackage {ulem}

\title{Interactive Medical Image Segmentation: A Benchmark Dataset and Baseline}

\author{
Junlong Cheng\textsuperscript{1,2,¶}
\qquad Bin Fu\textsuperscript{1}
\qquad Jin Ye\textsuperscript{1,3}
\qquad Guoan Wang\textsuperscript{1,4}
\qquad Tianbin Li\textsuperscript{1}
\\
\qquad \textbf{Haoyu Wang\textsuperscript{1,5}}
\qquad \textbf{Ruoyu Li\textsuperscript{2}}
\qquad \textbf{He Yao\textsuperscript{2}}
\qquad \textbf{Junren Chen\textsuperscript{2}}
\qquad \textbf{JingWen Li\textsuperscript{6}}
\\
\qquad \textbf{Yanzhou Su\textsuperscript{1}}
\qquad \textbf{Min Zhu\textsuperscript{2,§}}
\qquad \textbf{Junjun He\textsuperscript{1,‡}}
\\
\textsuperscript{1}Shanghai AI Laboratory, General Medical Artificial Intelligence\qquad \\ 
\textsuperscript{2}Sichuan University, School of Computer Science\qquad \\ 
\textsuperscript{3}Monash University\qquad \\ 
\textsuperscript{4}East China Normal University, School of computer science and technology\qquad \\ 
\textsuperscript{5}Shanghai Jiao Tong University, School of biomedical engineering\qquad \\ 
\textsuperscript{6}Xinjiang University, School of Computer Science and Technology\qquad \\ 
\textsuperscript{¶}Main technical contribution\qquad 
\textsuperscript{‡}Corresponding authors\qquad
\textsuperscript{§}Project lead\\ 
{\tt\small \{chengjunlong, yejin, hejunjun, litianbin\}@pjlab.org.cn}
}

\begin{document}

\maketitle

\begin{abstract}
Interactive Medical Image Segmentation (IMIS) has long been constrained by the limited availability of large-scale, diverse, and densely annotated datasets, which hinders model generalization and consistent evaluation across different models. In this paper, we introduce the IMed-361M benchmark dataset, a significant advancement in general IMIS research. First, we collect and standardize over 6.4 million medical images and their corresponding ground truth masks from multiple data sources. Then, leveraging the strong object recognition capabilities of a vision foundational model, we automatically generated dense interactive masks for each image and ensured their quality through rigorous quality control and granularity management. Unlike previous datasets, which are limited by specific modalities or sparse annotations, IMed-361M spans 14 modalities and 204 segmentation targets, totaling 361 million masks—an average of 56 masks per image. Finally, we developed an IMIS baseline network on this dataset that supports high-quality mask generation through interactive inputs, including clicks, bounding boxes, text prompts, and their combinations. We evaluate its performance on medical image segmentation tasks from multiple perspectives, demonstrating superior accuracy and scalability compared to existing interactive segmentation models. To facilitate research on foundational models in medical computer vision, we release the IMed-361M and model at \url{https://github.com/uni-medical/IMIS-Bench}.
\end{abstract}

\begin{figure}
    \centering
    \includegraphics[width=1\linewidth]{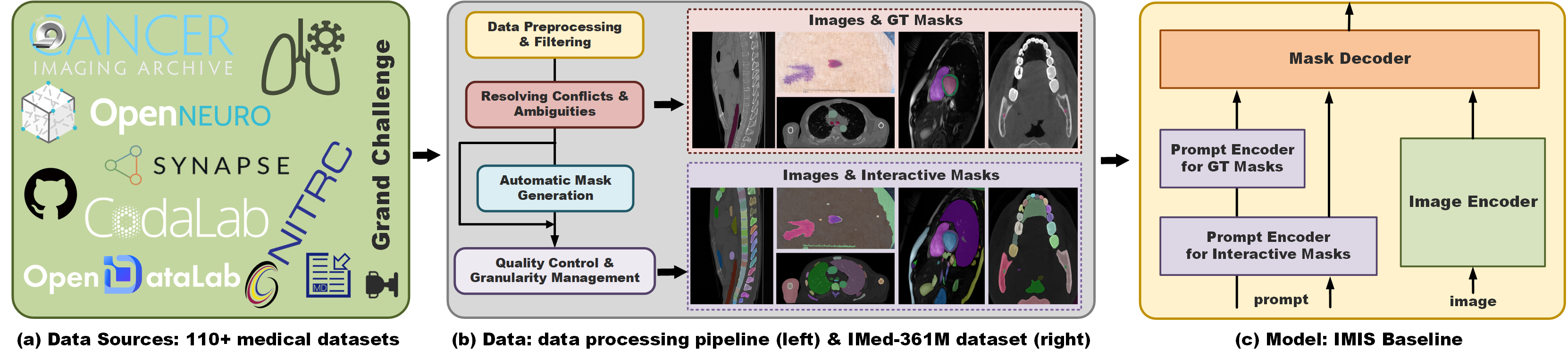}
    \caption{We collected 110 medical image datasets from various sources and generated the IMed-361M dataset, which contains over 361 million masks, through a rigorous and standardized data processing pipeline. Using this dataset, we developed the IMIS baseline network.}
    \label{fig1}
\end{figure}

\section{Introduction}

Interactive Medical Image Segmentation (IMIS) enables clinicians or users to guide the model by marking points, lines, or regions on an image, resulting in segmentation outcomes that better meet clinical needs~\cite{zhou2023volumetric,zhang2021interactive}.  
This user-informed segmentation approach not only optimizes results throughout the segmentation process, aiding clinicians in precisely localizing targets during diagnosis and treatment~\cite{chen2023joint,marinov2024deep,conze2023current}, but also addresses the limitations of fully automated segmentation models in generalizing to unseen object classes (i.e., zero-shot learning)~\cite{azad2024medical, cheng2022resganet,ronneberger2015u, chen2021transunet, zhou2018unet++, xing2024segmamba}, providing clinicians with a flexible adjustment mechanism that significantly enhances segmentation reliability. However, the advancement of IMIS advancement faces a critical bottleneck: the lack of high-quality, diverse, and large-scale datasets that capture the complex clinical environments~\cite{cascella2023evaluating, yuan2023devil}. Existing public datasets (such as Totalsegmentator~\cite{wasserthal2023totalsegmentator}, AutoPET~\cite{gatidis2022whole}, and CoNIC~\cite{graham2021conic}) are often tailored to specific modalities or tasks, thus constraining the generalization capabilities of models across diverse medical scenarios. 

Recently, as a foundational model for general visual segmentation, the Segment Anything Model (SAM) has garnered widespread attention~\cite{kirillov2023segment, ravi2024sam}. SAM integrates simple user interactions (such as points or bounding boxes) into the model’s learning process and leverages pre-training on large-scale datasets to achieve cross-domain and multi-task transferability~\cite{ke2024segment,cen2023segment,wang2024samrs}.
Although SAM has been pre-trained on over 1 billion natural image masks, its direct application in medical imaging remains limited due to the substantial domain differences between natural and medical images, as well as the lack of medical imaging-specific knowledge.
To address these issues, current methods such as MedSAM~\cite{ma2024segment} and SAM-Med2D~\cite{cheng2023sam} attempt to expand dataset size and fine-tune SAM to create IMIS models tailored for clinical scenarios. While these approaches have shown some progress, their ability to 'segment anything' in medical imaging remains limited. 
This limitation is mainly due to the lack of densely masks in existing medical datasets compared to the SA-1B dataset~\cite{kirillov2023segment} used to train SAM~\cite{ma2024segment,cheng2023sam,huang2024segment}. For instance, the COSMOS dataset~\cite{huang2024segment} contains an average of only 5.7 masks per image, which restricting the model's capacity for dense segmentation and hindering comprehensive, fine-grained interaction.
Moreover, many methods are evaluated only on specific modalities or with limited interaction strategies, further restricting the comprehensiveness and reliability of IMIS model evaluation~\cite{ma2024segment, zhang2024segment,huang2024segment, mazurowski2023segment}.

Addressing these limitations requires the development of a high-quality IMIS benchmark dataset, which is essential for advancing foundational models in medical imaging~\cite{marinov2024deep, zhang2024challenges,prabhod2024foundation,chen2024towards,schafer2024overcoming}. An ideal IMIS benchmark dataset should meet three core criteria: \textbf{(1) Large-scale.} The dataset should be large enough to fully support deep learning model training, enabling the model to effectively capture medical features; \textbf{(2) Diversity.} The dataset should encompass various medical imaging modalities and complex clinical scenarios to ensure the model’s ability to generalize across modalities and tasks; \textbf{(3) High-quality and densely masks.} Accurate segmentation of complex medical images relies on high-quality annotations, and densely masks further enhance the model’s capability to "segment anything" in medical imaging. Unfortunately, existing public medical segmentation datasets~\cite{ma2024segment,cheng2023sam,huang2024segment,xie2024medtrinity} do not fully meet these standards, limiting their ability to comprehensively support and evaluate IMIS models.

In this work, we introduce IMed-361M, a benchmark dataset specifically designed for IMIS tasks. As shown in Fig.\ref{fig1}, this dataset integrates public and private data sources and utilizes foundational models~\cite{kirillov2023segment} for automated annotation to generating dense masks for each image. A standardized data processing workflow is ensure the high quality and consistency of all masks. IMed-361M achieves unprecedented scale, diversity, and mask quality, comprising 6.4 million images spanning 14 imaging modalities and 204 targets, with a total of 361 million masks, averaging 56 masks per image. IMed-361M effectively solves the problems of small data size and sparse annotations, providing data support for IMIS model training.

Additionally, we develop an IMIS baseline model and conduct a comprehensive evaluation of its performance across various medical scenarios, including its effectiveness on different modalities, anatomical structures, and organs, as well as the impact of various interaction strategies on model outcomes. This analysis provided an in-depth understanding of the strengths and limitations of different interactive segmentation methods, establishing a fair and consistent framework for evaluating IMIS model performance.
We anticipate that the IMed-361M dataset and baseline model will drive the widespread adoption of IMIS technology in clinical practice, accelerating the healthcare industry's transition toward intelligence and automation.

\section{Related Work}

This section provides an overview of datasets used for various medical image segmentation tasks and reviews the advancements in IMIS algorithms.

\textbf{Datasets.} Due to variations in imaging protocols, medical images often have multidimensional characteristics (e.g., 2D, 3D), making pixel-level segmentation labeling for specific organs or lesions both expertise-dependent and time-consuming. Early work focused on single-organ or single-lesion annotated datasets~\cite{wasserthal2023totalsegmentator,gatidis2022whole,clark2013cancer,pedrosa2021lndb,rudyanto2014comparing,antonelli2022medical,jha2020kvasir}. These single-task datasets provide foundational data for training automatic segmentation models, significantly enhancing model performance in segmenting specific organs or lesions. With the growth of multi-task requirements, multi-organ and multi-tissue annotated datasets~\cite{wasserthal2023totalsegmentator,d2024totalsegmentator,zingman2024learning,podobnik2023han} have emerged, facilitating the exploration of deep learning models in multi-region segmentation tasks~\cite{isensee2021nnu,huang2023stu}. In recent years, visual foundational models 
have demonstrated impressive cross-domain and multi-task transfer capabilities, motivating researchers to build large-scale, multimodal datasets for medical foundational models. The most common approach is to integrate data from various sources~\cite{ma2024segment,cheng2023sam,huang2024segment,xie2024medtrinity} to bridge the scale gap compared to natural image datasets. However, due to the sparsity of annotations in the source data, the combined dataset still lags behind natural image datasets in terms of mask quantity and density. For example, the SA-1B~\cite{kirillov2023segment} dataset averages 100 masks per image, whereas large-scale medical datasets like COSMOS~\cite{huang2024segment} and SA-Med2D-20M~\cite{ye2023sa} have fewer than 5 masks per image on average. This gap limits the ability of medical datasets to support dense segmentation tasks and fine-grained interactions, as well as the applicability of foundational models to “segment anything” in the medical imaging domain.

\begin{figure*}[ht]
  \centering
  \includegraphics[width=1\textwidth]{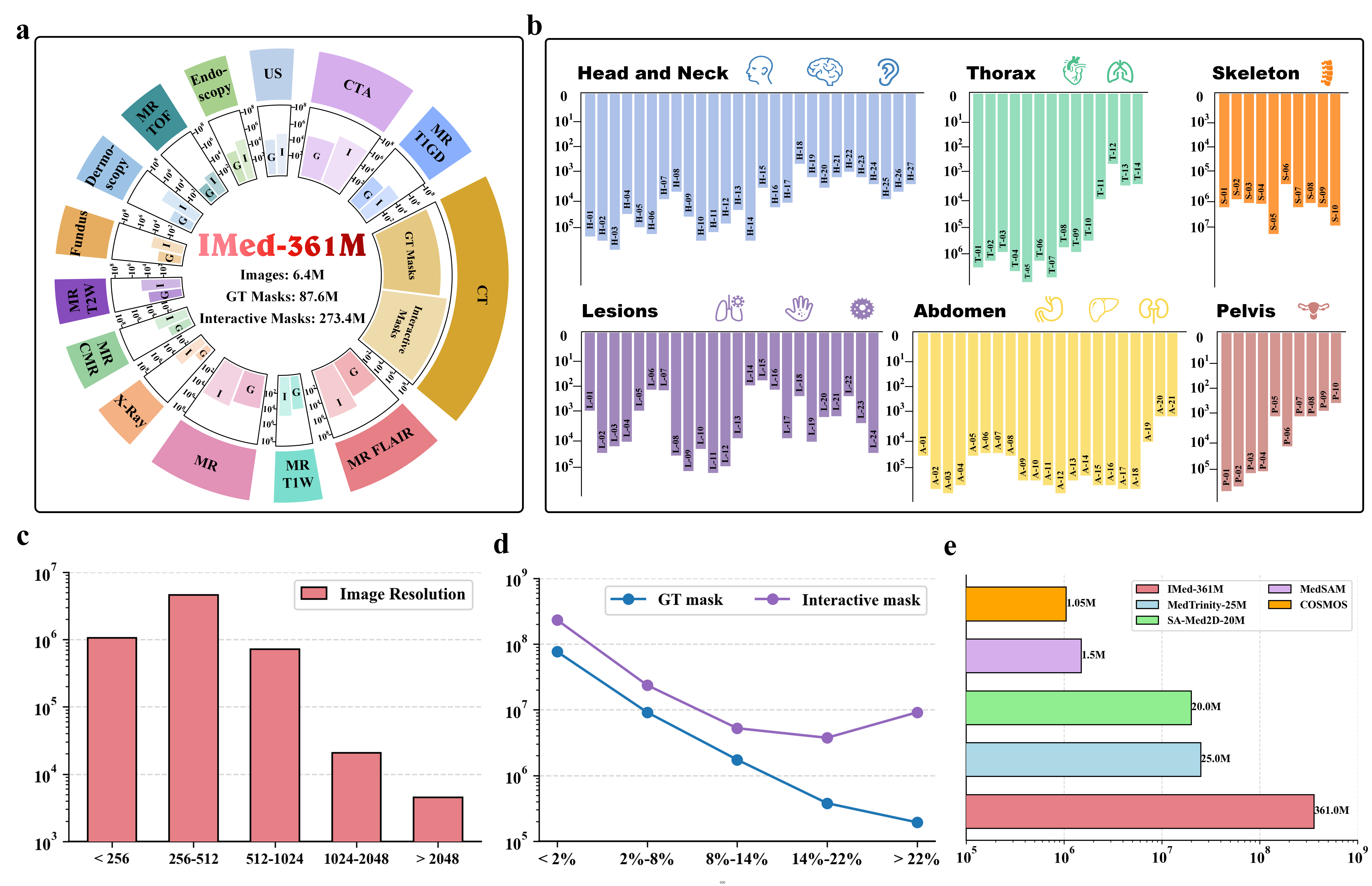}
  \caption{Overview of the IMed-361M dataset. (a) Number of images and masks for each modality. (b) Information on six anatomical structures. (c) Distribution of image resolutions. (d) Analysis of mask proportions. (e) Comparison with other existing public datasets.}\label{fig2}
\end{figure*}

\textbf{Algorithms.} IMIS methods are generally more effective than automated methods at generating high-quality results that meet clinical requirements, especially in scenarios involving diverse imaging protocols, complex pathological variations, and ambiguous lesion boundaries~\cite{olabarriaga2001interaction}. User interactions can be take several typical forms, such as scribbles~\cite{wong2023scribbleprompt,wang2016slic}, bounding boxes~\cite{kirillov2023segment,rajchl2016deepcut}, clicks~\cite{zhang2021interactive,sakinis2019interactive}, or language prompts~\cite{kirillov2023segment,du2023segvol, luddecke2022image}. Traditional methods approach this task through energy by minimizing energy on a regular pixel grid, capturing low-level appearance features with unary potentials and promoting consistent segmentation outputs with pairwise or higher-order potentials~\cite{wang2016slic,grady2005random}. With advancements in deep learning, researchers have explored using user prompts directly as network input features, often in combination with 2D or 3D neural networks to produce segmentation results~\cite{zhou2023volumetric,zhang2021interactive, kirillov2023segment, wong2023scribbleprompt, wang2016slic, sakinis2019interactive,rajchl2016deepcut}. Leveraging the advantage of pretraining on large-scale datasets, the Segment Anything Model (SAM) has become a benchmark for IMIS. For example, SAM can serve as a powerful pretrained encoder-decoder model fine-tuned for specific tasks~\cite{hu2024skinsam, li2024polyp}. However, these approaches essentially revert to the design of fully automated segmentation models. Another category of methods retains the interactive aspect of segmentation, focusing on parameter-efficient fine-tuning (PEFT) techniques~\cite{wu2023medical, paranjape2024adaptivesam}.
However, these methods consider only a limited of interaction strategies, and inconsistencies in evaluation have affected the comparability and reliability of their results. To achieve robust segmentation capabilities across diverse medical imaging datasets, some approaches fine-tune SAM's decoder using large-scale medical datasets~\cite{ma2024segment, cheng2023sam,huang2024segment}. Although these methods demonstrate significant advantages across different medical modalities and tasks, they have not yet been consistently benchmarked against similar methods and lack a thorough examination of how different interaction strategies affect outcomes.

In this work, we evaluate these state-of-the-art methods on the IMed-361M dataset, providing a comprehensive and fair comparison, while also exploring the impact of different interaction strategies on medical image segmentation.

\section{IMIS Benchmark Dataset: IMed-361M}

We present IMed-361M, the first large-scale IMIS dataset, that significantly surpasses existing datasets in terms of scale, diversity, quality, and density. This section details the data collection and pre-processing procedures, along with an in-depth analysis of the dataset's advantages and potential applications.

\subsection{Data Collection and Pre-processing}
\textbf{Data collection.} 
We integrate over 110 publicly available medical image segmentation datasets from globally recognized platforms such as TCIA\footnote{https://www.cancerimagingarchive.net}, OpenNeuro\footnote{https://openneuro.org}, NITRC\footnote{https://www.nitrc.org}, Grand Challenge\footnote{https://grand-challenge.org}, Synapse\footnote{https://www.synapse.org}, CodaLab\footnote{https://codalab.org}, and GitHub\footnote{https://github.com}, covering both 2D and 3D images and variety of formats (e.g., .jpg, .npy, .nii). Additionally, we collaborate with several medical institutions, acquiring diverse clinical data through rigorous ethical review processes, which further enriches the dataset. 
A detailed list of data sources is provided in the supplementary material.

\noindent \textbf{Preprocessing and filtering.}
We first standardize all collected medical images according to the SA-Med2D-20M~\cite{ye2023sa} protocol, converting the corresponding ground truth (GT) into one-hot encoding and storing it in .npz files in compressed sparse row (CSR) format~\cite{borvstnik2014sparse}. Then, we apply the following exclusion criteria: (1) exclude 3D slice images and their corresponding masks with an aspect ratio greater than 1.5 to avoid interference from images with excessive geometric distortion during model training; (2) exclude masks where the foreground area accounts for less than one-thousandth of the total pixels, as such small foreground regions are prone to being lost during resizing, making it challenging to generate effective prompts. 

\noindent \textbf{Resolving conflicts and ambiguities.} 
To address conflicts and ambiguities in the GT, we first standardize expressions for the same target. For instance, both "lung nodule" and "pulmonary nodule" are renamed to "lung nodule" for consistency. We then manually review and correct misalignments and informational errors in the dataset to ensure annotation accuracy. Finally, for annotations with multiple connected components, we differentiate and label them based on clinical needs to avoid potential misunderstandings that could arise from single-point interactions.

Through this process, we collected 6.4 million images and 87.6 million GT masks that were manually annotated and corrected. Although this already exceeds the data volume of most existing segmentation datasets, it still does not meet our fundamental requirement for the IMIS dataset: diverse and densely masks. Therefore, we further introduce interactive masks to address the above issues.

\subsection{Interactive Masks}

\textbf{Automatic mask generation.} The automatic mask generation method employed by SAM has been proven to be high-quality and effective~\cite{kirillov2023segment,he2024weakly, zhang2023input}. We leverage SAM's object-awareness capability to generate as many masks as possible for each image. Specifically, we employ a 32×32 point grid to guide the model, generating a set of candidate masks for each point corresponding to potential objects of interest. The generated masks are refined and optimized using the following strategies: (1) Confidence filtering: Only masks with an Intersection over Union (IoU) prediction confidence score above 0.85 are retained to ensure segmentation accuracy. (2) Non-Maximum Suppression (NMS)~\cite{symeonidis2023neural}: For overlapping masks with an IoU greater than 0.7, only the mask with the highest confidence score is kept to eliminate redundancy. 
(3) Remove background masks: Masks covering more than 80\% of the area are discarded, as such large-coverage foreground masks are almost nonexistent in the GT.

\noindent \textbf{Quality control and granularity management.} Based on our observations, the generated masks often fail to fully separate structures with unclear boundaries, such as the left atrium, myocardium, and right atrium in the heart. Additionally, masks for dispersed structures like the intestines are often identified as multiple separate objects. To address these issues, we utilize the original GT (described in Section 3.1) to correct the generated masks. We focus on two situations: (1) If the GT contains multi-connected regions, we directly replace the corresponding regions in the generated mask with the multi-connected regions from the GT; (2) We iterate over all single-connected regions in the GT, generating a minimum bounding box for each region. If there is a region in the generated mask with a bounding box that overlaps more than 95\% with this GT bounding box, we retain the generated mask region; otherwise, we replace the region with the GT mask. Finally,  
we apply morphological operations (e.g., erosion and dilation) to remove noise and fill small holes~\cite{tummala2023morphological}.

We ultimately obtained 273 million "interactive masks", which can be used to train interactive segmentation models and cover nearly all identifiable objects in medical images. While we acknowledge that some masks may lack direct clinical relevance, they provide considerable diversity and density in the identified regions, helping the model learn a variety of interactive operations and enhancing its generalization ability across various tasks.

\begin{figure}[ht]
  \centering
  \includegraphics[width=0.8\textwidth]{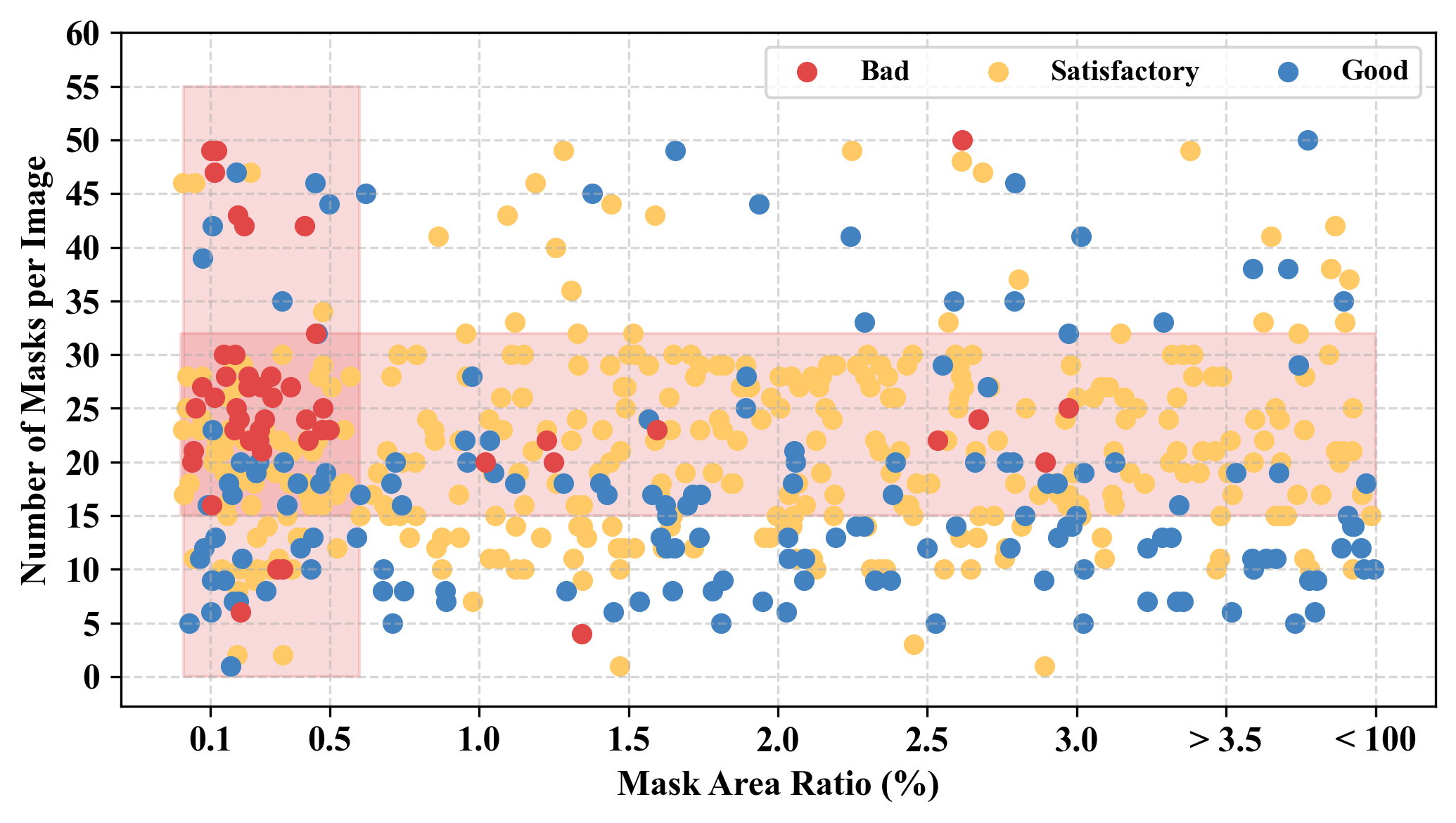}
  \caption{Evaluation of the quality of interactive masks.}\label{fig3}
\end{figure}

\subsection{Statistics and Analysis}

\textbf{Data scale.} As shown in Fig.\ref{fig2} (a), the IMed-361M dataset contains 6.4 million images, 87.6 million GT, and 273.4 million interactive masks, averaging 56 masks per image. This makes it the largest publicly available, multimodal, interactive medical image segmentation dataset to date. Compared to the MedTrinity-25M~\cite{xie2024medtrinity} dataset, our dataset provides 14.4 times the number of masks (Fig.\ref{fig2} (e)).

\noindent \textbf{Diversity.} The IMed-361M dataset covers 14 imaging modalities and 204 segmentation targets, including organs and lesions, as shown in Fig.\ref{fig2} (b). We categorize the GT into six groups: Head and Neck, Thorax, Skeleton, Abdomen, Pelvis, and Lesions, covering nearly all parts of the human body. For clarity, we merge certain positional information when appropriate; for example, “left lung” and “right lung” are combined into “lung.” Detailed category information can be found in the supplementary materials. Fig.\ref{fig2} (c) and (d) show the distribution of image resolutions and mask coverage within the dataset. Over 83\% of the images have resolutions between 256×256 and 1024×1024, ensuring broad applicability and compatibility across various research scenarios. Additionally, most masks occupy less than 2\% of the image area, reflecting the typically fine granularity of medical segmentation. Our interactive masks provide over one million instances across different coverage intervals, significantly enhancing the diversity and density of the dataset.

\noindent \textbf{Mask quality.} We randomly select five images from each subset of IMed-361M (a total of 550 images) and invited four radiologists to evaluate the quality of the corresponding interactive masks. We categorize the mask quality into three levels: (1) "Good": the mask closely matches the manual annotations; (2) "Satisfactory": the mask has some visible errors (e.g., incomplete contours) but is generally acceptable; (3) "Bad": the mask has significant errors (e.g., over-segmentation or under-segmentation) and is unsuitable for clinical segmentation tasks. The evaluation results show that 156 images are rated as "Good," 345 as "Satisfactory," and 49 as "Bad". Statistical analysis reveals that the masks rated as "Bad" mainly come from 18 datasets, concentrated within the red-boxed area in Fig.\ref{fig3}. Radiologists note that some excessively annotated structures hold little clinical relevance. Additionally, some masks include elements unrelated to human organs or lesions, such as letters in X-ray backgrounds or hair in dermatoscopic images. Based on this analysis, we manually remove the excessively annotated masks from these 18 datasets and apply a 0.5\% foreground filter rate to retain only the final valid masks. Finally, we retain those masks unrelated to human organs or lesions to enhance the model's adaptability in different scenarios.

\section{IMIS Baseline Network}

\subsection{Model Design}

\begin{figure}[t]
  \centering
  \includegraphics[width=1 \textwidth]{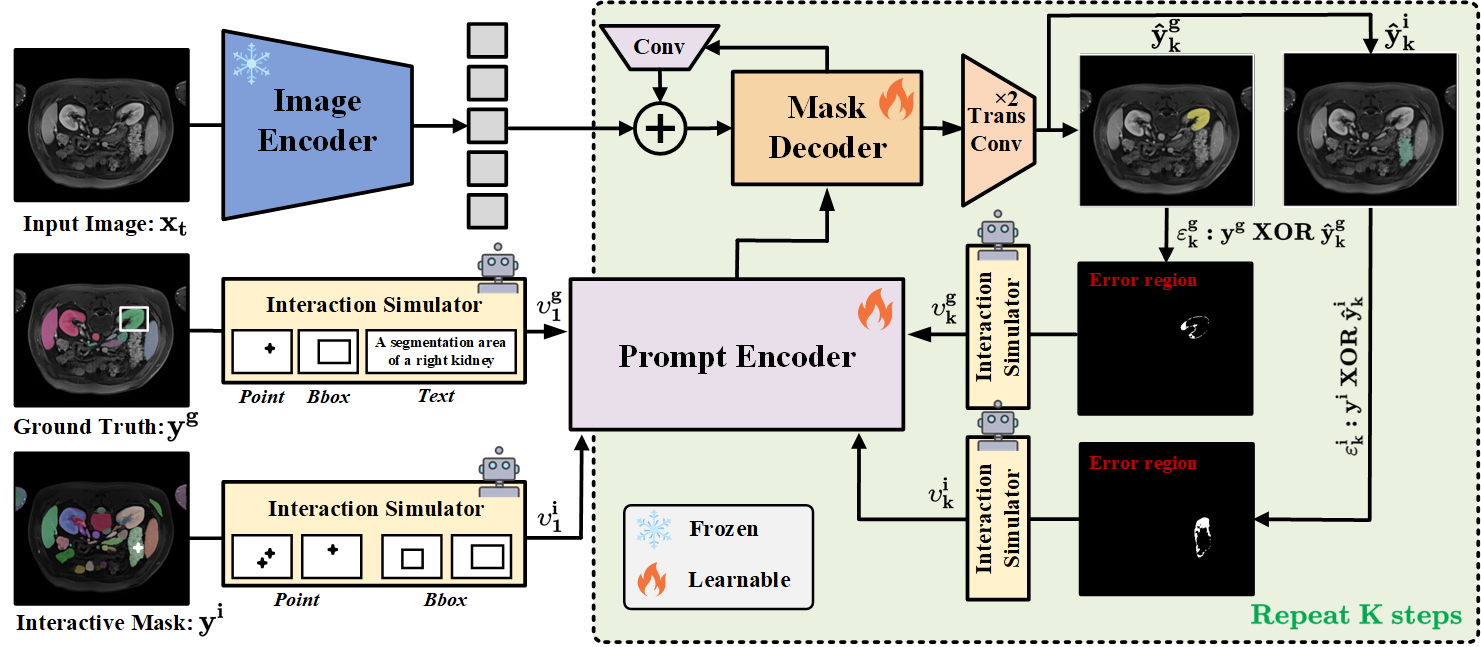}
  \caption{The training process of IMIS-Net simulates K consecutive steps of interactive segmentation.}\label{fig4}
\end{figure}

We adopt a strategy similar to fine-tuning SAM~\cite{ma2024segment,cheng2023sam,huang2024segment} to establish the IMIS baseline, providing a performance benchmark for future work. As shown in Fig.\ref{fig4}, IMIS-Net has three key components: an image encoder for extracting image features, a prompt encoder for integrating user interaction information, and a mask decoder for generating segmentation results using image and prompt embeddings.

We select ViT-base~\cite{dosovitskiy2020image} as the image encoder. While larger ViT models (e.g., ViT-Large, ViT-Huge)~\cite{ma2024segment,huang2024segment} offer slight accuracy improvements, they significantly increase computational costs, making them impractical for clinical applications. Additionally, most open-source fine-tuning methods also use ViT-base, ensuring fairness in our baseline. The image input size is 1024 × 1024 × 3, with a patch size of 16 × 16 × 3. For the prompt encoder, we consider three prompt types: points, boxes, and text. Points and boxes are represented by the sum of positional encoding and learned embeddings, while text is encoded using CLIP's text encoder~\cite{radford2021learning}. Text prompts follow the template: "A segmentation area of a $[category]$," covering over 200 organ and lesion categories to establish a benchmark for future multimodal segmentation research. The mask decoder uses Transformer decoder blocks, mapping image and prompt embeddings to the mask. The mask resolution is increased to 256 × 256 through transposed convolutions and matched to the input size via bilinear interpolation. Notably, expanding the encoder can improve performance without significantly increasing network training parameters, thereby providing a simple solution to address the issue of model performance saturation.

\subsection{Experiment Configuration}

\textbf{Training strategy.} 

An overview is provided of the simulated continuous interactive segmentation training~\cite{ma2024segment,cheng2023sam,sofiiuk2022reviving} in Fig.\ref{fig4}. For a given segmentation task and medical image $x_t$, we first simulate a set of initial interactions $u^{g}_{1}$ and $u^{i}_{1}$ based on the corresponding ground truth $y^g$ and interactive mask $y^i$, which include clicks, bboxes, and text input. The click points are uniformly sampled from the foreground regions of $y^g$ or $y^i$, while the bboxes are defined as the smallest bounding box around the target, with an offset of 5 pixels added to each coordinate to simulate slight user bias during the interaction process. The entire training process involves $K$ interactive training iterations (with $K=8$ in this paper). The model's initial predictions are $\hat{y}^{g}_{1}$ and $\hat{y}^{i}_{1}$. After the first prediction, we simulate subsequent corrections based on the previous predictions $\hat{y}^{g}_{k}$ and $\hat{y}^{i}_{k}$, as well as the error region $\varepsilon_{k}$ between the $y^g$ and $y^i$, where $k\in \{1,...,K\}$. Additionally, we provide the low-resolution predicted mask from the previous prediction as an extra cue to the model.  As can be seen, the image encoder only needs to encode the image once during the training, and subsequent interactive training only updates the prompt encoder and mask decoder parameters.

\noindent \textbf{Training setting.} We use the Adam optimizer~\cite{zhang2018improved} with a learning rate of $2\times10^{-5}$. The training was conducted on 72 NVIDIA 4090 GPUs with a batch size of 2. For each image, we randomly select 5 targets from the corresponding GT and interactive masks as supervision targets (if fewer than 5 targets are present, they are selected repeatedly). The image resolution was uniformly resized to $1024\times1024$, and pixel intensities are randomly scaled and shifted with a probability of 20\%, where the scaling factor and offset are set to 0.2. We train the model for a total of 12 epochs on the IMed-361M dataset and select the final checkpoint as the model weights. A linear combination of Focal loss~\cite{ross2017focal} and Dice loss~\cite{milletari2016v} is used as the loss function, balancing their influence in a 20:1 ratio. Finally, the Dice score is used as the primary evaluation metric for this study.

\begin{figure}[h]
  \centering
  \includegraphics[width=0.95 \textwidth]{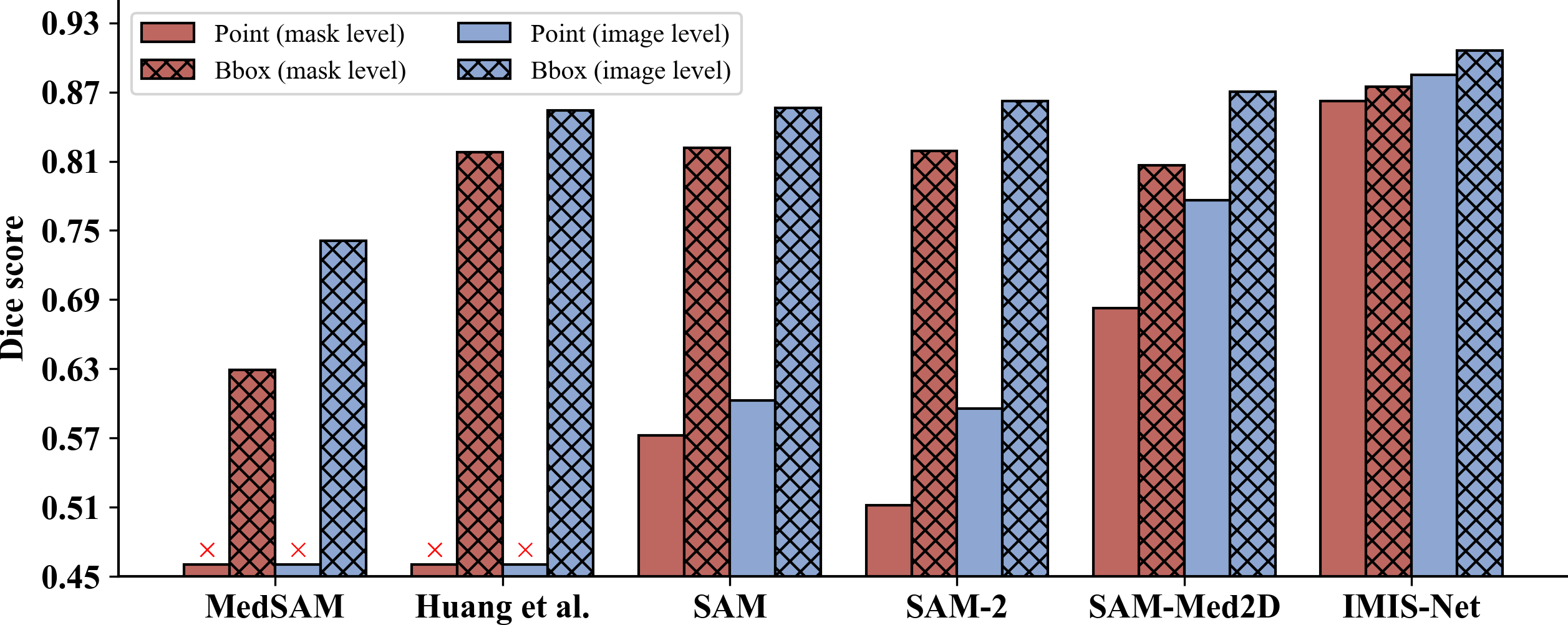}
  \caption{Comparison of IMIS-Net with existing foundation models, with performance statistics at both image and mask levels.}\label{fig5}
\end{figure}

\subsection{Evaluation Model and Data}

\textbf{Model.} We evaluate five visual foundation models~\cite{kirillov2023segment,ravi2024sam,ma2024segment,cheng2023sam,huang2024segment}, selected based on their pretraining on large-scale datasets and the availability of source code. Among them, MedSAM, SAM-Med2D, and Huang et al.~\cite{huang2024segment} are specifically designed for IMIS, while SAM~\cite{kirillov2023segment} and SAM-2~\cite{ravi2024sam} are pre-trained on massive natural image and video datasets. To ensure fairness in the evaluation, all models used ViT-Base as the encoder. It is important to note that, MedSAM and Huang et al.~\cite{huang2024segment} only support bbox-based interactivity (using the models released by the authors), while the other models support points, bboxes, and combinations of both.

\noindent \textbf{Data.} The data used to evaluate the model is divided into two categories: the test set and the external dataset. The test set contains 58,411 images, with a distribution consistent with that of IMed-361M. The external dataset includes data from three different modalities and clinical scenarios: SegThor~\cite{lambert2020segthor}, which is used for segmentation of thoracic organs at risk in 516 CT images; TotalSegmentator MRI~\cite{d2024totalsegmentator}, which is used for segmenting organs in 2,147 MRI images; and ISLES~\footnote{https://www.isles-challenge.org/}, which belongs to the MRI FLAIR modality and includes 364 images of ischemic stroke lesions. These datasets were not used during the training of IMIS-Net.

\begin{figure*}[t]
  \centering
  \includegraphics[width=0.95 \textwidth]{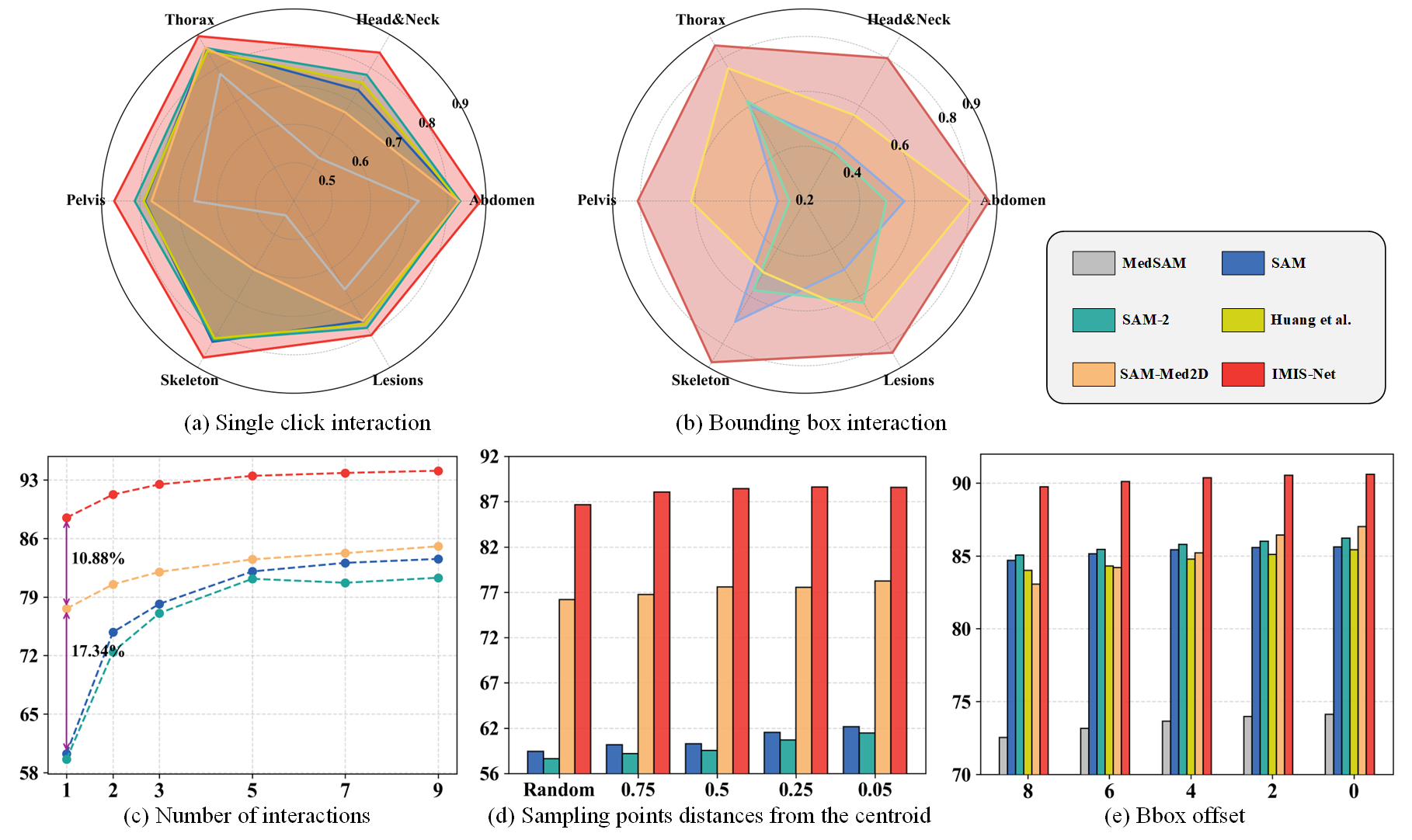}
  \caption{Comparison of segmentation performance across different anatomical structures under single-click (a) and bounding box (b) interactions. (c) Changes in segmentation performance with increasing interaction numbers. (d) and (e) Impact of click position and bounding box offset on performance.}\label{fig6}
\end{figure*}

\begin{table*}[ht]
\centering

\resizebox{1\textwidth}{!}{
\begin{tabular}{llcccccc}
\hline
\toprule
Dataset                                                                          & Category           & SAM    & SAM 2           & MedSAM & Huang et al.    & SAM-Med2D       & IMIS-Net             \\ \hline
\toprule
ISLES                                                                   & Ischemic Stroke Lesion        & 55.92 & 60.14 & 59.90 & 56.79          & 68.22          & \textbf{71.78}          \\ \hline
\multirow{5}{*}{SegThor}                                                         & Esophagus          & 78.19 & 86.60          & 47.17 & 76.48          & 82.04          & \textbf{89.17} \\
                                                                                 & Heart              & 91.41 & 92.34          & 81.16 & 92.02          & \textbf{94.18} & 81.27          \\
                                                                                 & Aorta              & 82.36 & 83.69          & 55.68 & 84.57          & 79.12          & \textbf{94.92} \\
                                                                                 & Trachea            & 93.76 & 94.24          & 72.40 & \textbf{94.60} & 85.10          & 90.04          \\ \cline{2-8} 
                                                                                 & \textbf{Average}   & 84.46 & 85.86          & 60.55 & 84.73          & 86.43          & \textbf{89.27} \\ \hline
\multirow{16}{*}{\begin{tabular}[c]{@{}l@{}}Totalsegmentator\\ MRI\end{tabular}} & Adrenal gland      & 77.19 & \textbf{82.25} & 56.58 & 68.52          & 66.35          & 77.53          \\
                                                                                 & Aorta              & 76.88 & 79.90          & 55.98 & 80.08          & 74.91          & \textbf{85.62} \\
                                                                                 & Colon              & 56.69 & 56.00          & 42.53 & 55.41          & 51.44          & \textbf{73.47} \\
                                                                                 & Duodenum           & 69.30 & 74.81          & 55.64 & 72.38          & 69.67          & \textbf{75.97} \\
                                                                                 & Gallbladder        & 81.33 & 84.82          & 68.15 & 83.46          & \textbf{87.63} & 82.93         \\
                                                                                 & Iliopsoas          & 59.57 & 71.96          & 58.48 & 63.67          & 60.76          & \textbf{72.55} \\
                                                                                 & Kidney             & 79.43 & 81.59         & 58.41 & 81.09         & 74.04            & \textbf{82.72}          \\
                                                                                 & Inferior vena cava & 84.94 & 87.03          & 57.73 & 85.88          & 78.34          & \textbf{87.35} \\
                                                                                 & Liver              & 86.93 & 88.70          & 68.14 & 87.35          & 86.75          & \textbf{90.62} \\
                                                                                 & Pancreas           & 66.95 & 69.86          & 46.10 & 66.30          & 66.84          & \textbf{70.84} \\
                                                                                 & Spleen             & 81.12 & 83.05          & 66.61 & 83.87          & 84.95          & \textbf{86.74} \\
                                                                                 & Stomach            & 77.10 & 81.23          & 62.75 & 76.52          & 79.31          & \textbf{81.45} \\ \cline{2-8} 
                                                                                 & \textbf{Average}   & 75.45 & 77.62          & 59.52 & 75.47          & 75.92          & \textbf{79.06} \\ \hline
\toprule
\end{tabular}
}
\caption{Quantitative comparison results of IMIS-Net against five other interactive segmentation methods on three external datasets.}\label{tab1}
\end{table*}

\begin{figure*}[ht]
  \centering
  \includegraphics[width=1\textwidth]{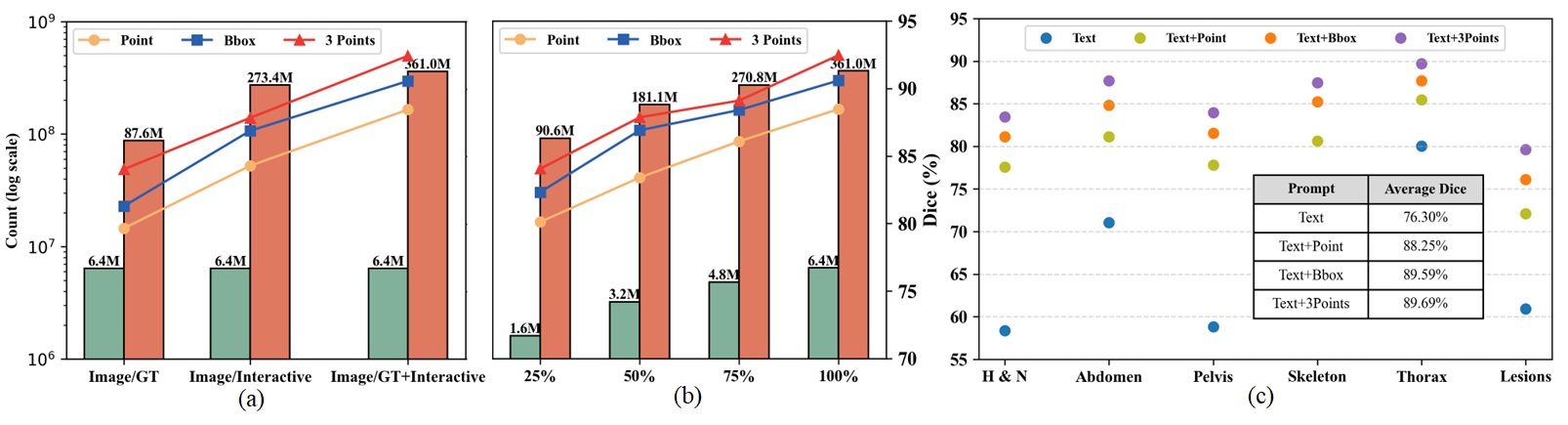}
  \caption{(a) and (b) segmentation performance increases with the increasing mask density or data scales up. (c) effectiveness of text and its combination with other prompts.}\label{fig7}
\end{figure*}

\section{Results}

\subsection{Main Results} 
Fig.\ref{fig5} shows the performance evaluation of IMIS-Net compared to other vision foundation models on the single-interaction segmentation task. During testing, all models used the same points and bounding boxes as prompts. The results indicate that IMIS-Net outperforms other models in both image and mask-level statistics. The bbox interaction of different models always outperforms the click interaction because the bbox can provide more boundary information. Notably, despite being pretrained on large-scale medical image datasets, MedSAM and SAM-Med2D still exhibit significant performance differences. This disparity is closely related to the scale and diversity of the pretraining datasets. As shown in Fig.\ref{fig6}(a) and (b), the pre-training dataset for SAM-Med2D lacks samples of skeletal structures, resulting in poorer segmentation performance for these anatomical structures. Additionally, under the single-point prompt condition, SAM and SAM-2 achieve only Dice scores of 60.26\% and 59.57\%, respectively. likely due to the absence of medical knowledge in the pretraining data and limited interactive information constraining model performance.

We increase the number of interactions from 1 to 9 to observe the performance changes of the model (Fig.\ref{fig6} (c)). As expected, performance improves with more interactions, and the gap between models narrows. This is because multiple interactions reduce the task difficulty. We also examine the impact of click position and bbox offset on performance. As shown in Fig.\ref{fig6} (d) and (f), when the prompt point is closer to the centroid, the performance improvement is more significant, with SAM-2’s Dice score increasing by 2.84\%. Additionally, with bbox offset, all methods experience a performance decline of 0.85\%-3.94\%. Our model exhibited the smallest performance drop, making it more robust for practical use. In summary, leveraging the rich data and mask diversity of the IMed-361M dataset, our model delivers the best performance across various medical scenarios and interaction strategies. More experimental results and analysis can be found in the supplementary material.

\vspace{-0.5em}
\subsection{External Dataset Evaluation}
\vspace{-1em}
Tab.\ref{tab1} presents evaluation results on external datasets. IMIS-Net achieves the best average performance across datasets from three different sources and tasks. We highlight the Dice scores of the 12 major abdominal organs in the TotalSegmentator MRI dataset, where IMIS-Net outperforms in 10. On the ISLES dataset, focusing on ischemic stroke lesion segmentation, our model surpasses the second-best SAM-Med2D by 3.56\%. These results demonstrate the strong generalization ability of our method.


\begin{table}[h]    
\centering    

\setlength{\tabcolsep}{1.3pt} 
\renewcommand{\arraystretch}{1.2} 
\begin{tabular}{>{\centering\arraybackslash}m{1.5cm} >{\centering\arraybackslash}m{1.5cm} >{\centering\arraybackslash}m{1.5cm} >{\centering\arraybackslash}m{1.5cm} >{\centering\arraybackslash}m{2cm}}    
\hline    
\toprule  
\multirow{2}{*}{\begin{tabular}[c]{@{}c@{}}Decoder\\ dimension\end{tabular}} & \multirow{2}{*}{\begin{tabular}[c]{@{}c@{}}Image\\ resolution\end{tabular}} & \multicolumn{2}{c}{Prompt} & \multirow{2}{*}{\begin{tabular}[c]{@{}c@{}}Training \\ parameters\end{tabular}} \\ \cline{3-4}  
 & & Point & Bbox &  \\ \hline    
\toprule  
768 & 256×256 & 0.8214 & 0.8469 & 29.68 M \\    
768 & 512×512 & 0.8673 & 0.8968 & 29.68 M \\    
256 & 1024×1024 & 0.8366 & 0.8497 & 5.52 M \\    
512 & 1024×1024 & 0.8563 & 0.8729 & 15.19 M \\    
768 & 1024×1024 & 0.8848 & 0.9060 & 29.68 M \\ \hline  
\toprule  
\end{tabular}    
\caption{Ablation study of model design, including decoder dimension and image resolution.}\label{tab2}
\end{table}

\subsection{Ablation Study} 

\textbf{Scaling Up Training Data.} 
As shown in Fig.\ref{fig4} (a) and (b), when trained only on the GT from IMed-361M, IMIS-Net performs poorly. However, with the addition of interactive masks, the Dice score increases rapidly. Similar performance trends are observed when training on datasets of different sizes, indicating that our method is scalable and performs better with more available training data.

\noindent \textbf{Text and Combination Prompts.} 
IMIS-Net, as a versatile interactive model, can utilize text prompts to achieve the segmentation of over 200 organs and lesions. As shown in Fig.\ref{fig4} (c), the model achieves a segmentation performance of 76.30\% when using only text prompts. When both text and point prompts are combined, the average Dice score increases by 11.95\%. Furthermore, after three rounds of click-based correction, the Dice score reaches 89.69\%. These results show that combining different prompts synergistically enhances the model’s segmentation capabilities.

\noindent \textbf{Model Design.} 
We conduct an ablation study on different configurations of the model, including the impact of input resolution and decoder dimension on model performance. As shown in Tab.\ref{tab2}, the model’s segmentation performance improves as the training image resolution increases. This higher input resolution allows for clearer visualization of lesions and organs in medical images. Additionally, when the decoder dimension is increased from 256 to 768, the model's performance improves from 84.97\% to 90.60\%, with only a 24.16M increase in trainable parameters. These results demonstrate the scalability of IMIS-Net, showing that even when the model's performance approaches saturation on larger datasets, further performance improvements can still be achieved by expanding the decoder.

\section{Conclusion}
In this work, we introduce IMed-361M, a benchmark dataset dedicated to interactive medical image segmentation. It includes a vast array of medical images across various modalities, extensive segmentation scenarios, and densely masks, surpassing all existing datasets that are limited to single tasks or simple integrations. Leveraging this data resource, we developed a general IMIS baseline model that enables users to generate segmentation results tailored to clinical needs through interactive methods, including clicks, bounding boxes, text prompts, and their combinations. We performed a comprehensive comparison of this baseline with existing foundational models, demonstrating that our model provides significant performance advantages and exhibits strong transferability in previously unseen scenarios. Notably, our approach typically requires fewer interactions to achieve comparable performance, enhancing its practicality in real-world applications.

The IMed-361M dataset will strongly facilitate the development of foundational models in medical imaging and lay the foundation for fair evaluation across different models. IMIS-Net provides general technical support for various clinical applications, accelerating the widespread application of AI technology in the medical field. Despite these achievements, we recognize that this work still faces several challenges. For instance, effectively obtaining semantic information for interactive masks and extending this approach to more comprehensive and finer-grained medical image analysis scenarios are areas that require further exploration and improvement in the future.


%

\maketitle












\section*{Appendix A: Demo and Code}

Our code, model weights, and dataset are available at: \url{https://github.com/uni-medical/IMIS-Bench}.

\begin{figure}[h]
    \centering
    \includegraphics[width=0.95\linewidth]{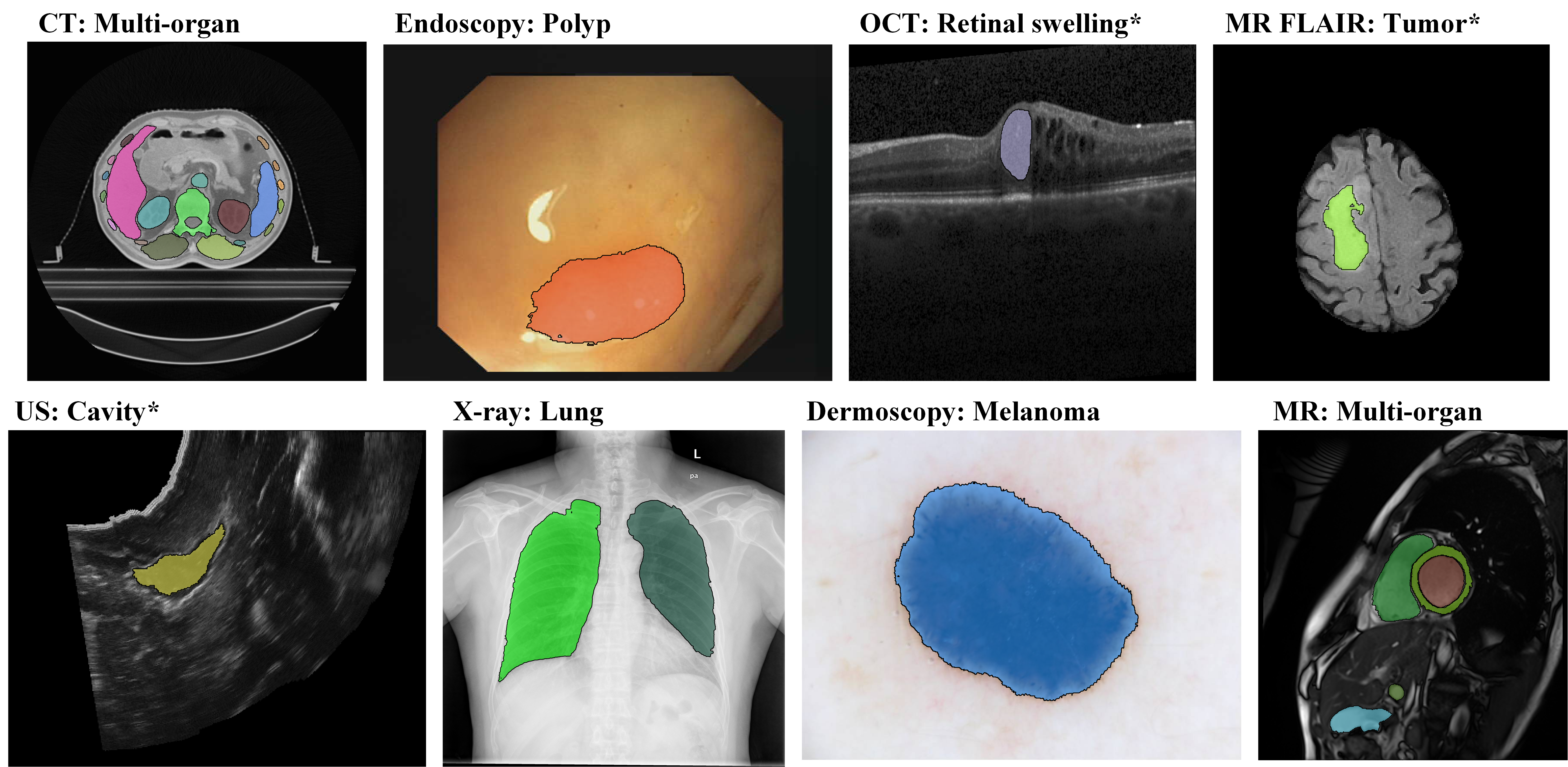}
    \caption{\textbf{Example predictions of IMIS-Net across different modalities and segmentation tasks}. "*" Indicates that the corresponding image modality or segmentation task was not included in our training plan. Our model demonstrates its versatility by effectively handling multiple medical image modalities and performing various segmentation tasks, even on those that it has not previously encountered.}
    \label{sup_fig1}
\end{figure}

\section*{Appendix B: IMed-361M Information and Availability}

\begin{figure}[h]
    \centering
    \includegraphics[width=1\linewidth]{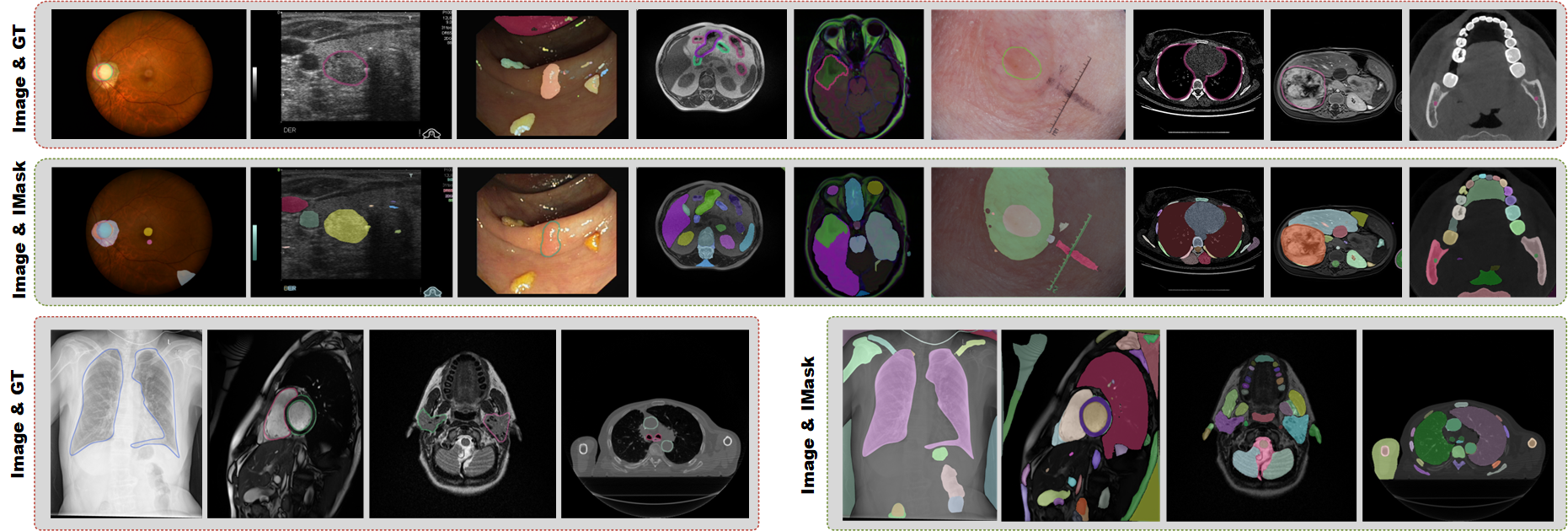}
    \caption{\textbf{IMed-361M:} A comprehensive dataset of multimodal medical images encompassing nearly all human organs and lesions, with interactive masks offering detailed, dense annotations.}
    \label{sup_fig2}
\end{figure}

We have compiled 110 medical image segmentation datasets into a comprehensive, large-scale, multimodal, high-quality dataset named IMed-361M, making it openly accessible for interactive medical image segmentation. This dataset includes over 6.4 million images, 87.6 million ground truth annotations, and 273.4 million interactive masks (IMask), encompassing 14 image modalities and 204 segmentation targets. Fig.\ref{sup_fig2} presents representative samples, while detailed category information is provided in Tab.\ref{sup_tab1}, which forms the basis for IMIS-Net's text prompts. The modality and category details of these open-source and private datasets are displayed in Tab.\ref{sup_tab2}.

\definecolor{acolor}{HTML}{FBE076}  
\definecolor{hcolor}{HTML}{ADC1E9}  
\definecolor{scolor}{HTML}{FE993E}  
\definecolor{tcolor}{HTML}{95DDBC}  
\definecolor{pcolor}{HTML}{D29590}  
\definecolor{lcolor}{HTML}{A088BC}  

\begin{table}[h]
    \centering
    \scriptsize
    \renewcommand{\arraystretch}{1.3} 
    \caption{IMed-361M dataset contains six anatomical categories: A (Abdomen), S (Skeleton), H (Head \& Neck), T (Thorax), P (Pelvis), and L (Lesions). The symbol $\ast$ indicates that the target has left and right parts.}
    
    \begin{tabular}{p{3cm} p{3cm} p{3cm} p{3cm}}  
  
        \midrule
        \cellcolor{acolor}A01: Adrenal gland $\ast$ & \cellcolor{hcolor}H07: Optic chiasm & \cellcolor{scolor}S07: Scapula $\ast$ & \cellcolor{lcolor}L01: Lung infections \\
        \cellcolor{acolor}A02: Aorta & \cellcolor{hcolor}H08: Pituitary gland & \cellcolor{scolor}S08: Cervical spine (C1-C7) & \cellcolor{lcolor}L02: Liver tumor \\
        \cellcolor{acolor}A03: Autochthonous muscles$\ast$ & \cellcolor{hcolor}H09: Brain stem & \cellcolor{scolor}S09: Lumbar spine (L1-L6) & \cellcolor{lcolor}L03: Kidney tumor \\
        \cellcolor{acolor}A04: Colon & \cellcolor{hcolor}H10: Temporal lobe $\ast$ & \cellcolor{scolor}S10: Thoracic spine (T1-T13) & \cellcolor{lcolor}L04: Kidney cyst \\
        \cellcolor{acolor}A05: Duodenum & \cellcolor{hcolor}H11: Parotid gland $\ast$ & \cellcolor{tcolor}T01: Esophagus & \cellcolor{lcolor}L05: Pleural effusion \\
        \cellcolor{acolor}A06: Gallbladder & \cellcolor{hcolor}H12: Ear ($\ast$, Inner, Middle) & \cellcolor{tcolor}T02: Atrium $\ast$ & \cellcolor{lcolor}L06: Myocardial edema \\
        \cellcolor{acolor}A07: Iliac artery $\ast$ & \cellcolor{hcolor}H13: Temporomandibular $\ast$ & \cellcolor{tcolor}T03: Myocardium $\ast$ & \cellcolor{lcolor}L07: Myocardial scars \\
        \cellcolor{acolor}A08: Iliac vein $\ast$ & \cellcolor{hcolor}H14: Mandible $\ast$ & \cellcolor{tcolor}T04: Ventricle $\ast$ & \cellcolor{lcolor}L08: Necrosis \\
        \cellcolor{acolor}A09: Iliopsoas $\ast$ & \cellcolor{hcolor}H15: Thyroid gland & \cellcolor{tcolor}T05: Lower lobe $\ast$ & \cellcolor{lcolor}L09: Edema \\
        \cellcolor{acolor}A10: Inferior vena cava & \cellcolor{hcolor}H16: Submandibular gland $\ast$ & \cellcolor{tcolor}T06: Middle lobe $\ast$ & \cellcolor{lcolor}L10: Non enhancing tumor \\
        \cellcolor{acolor}A11: Kidney $\ast$ & \cellcolor{hcolor}H17: Oral cavity & \cellcolor{tcolor}T07: Upper lobe $\ast$ & \cellcolor{lcolor}L11: Enhancing tumor \\
        \cellcolor{acolor}A12: Liver & \cellcolor{hcolor}H18: Eustachian tube $\ast$ & \cellcolor{tcolor}T08: Pulmonary artery & \cellcolor{lcolor}L12: Necrotic tumor core \\
        \cellcolor{acolor}A13: Pancreas & \cellcolor{hcolor}H19: Hippocampus $\ast$ & \cellcolor{tcolor}T9: Trachea & \cellcolor{lcolor}L13: Peritumoral edema \\
        \cellcolor{acolor}A14: Portal and splenic veins & \cellcolor{hcolor}H20: Mastoid $\ast$ & \cellcolor{tcolor}T10: Lung & \cellcolor{lcolor}L14: Myocardial infarction \\
        \cellcolor{acolor}A15: Small intestine & \cellcolor{hcolor}H21: Tympanic cavity $\ast$ & \cellcolor{tcolor}T11: Heart & \cellcolor{lcolor}L15: No reflow \\
        \cellcolor{acolor}A16: Spleen & \cellcolor{hcolor}H22: Semicircular canal $\ast$ & \cellcolor{tcolor}T12: Bronchus $\ast$ & \cellcolor{lcolor}L16: Brain aneurysm \\
        \cellcolor{acolor}A17: Stomach & \cellcolor{hcolor}H23: Optic cup & \cellcolor{tcolor}T13: Breast $\ast$ & \cellcolor{lcolor}L17: Neuroblastoma \\
        \cellcolor{acolor}A18: Spinal cord & \cellcolor{hcolor}H24: Optic disc & \cellcolor{tcolor}T14: Ascending aorta & \cellcolor{lcolor}L18: Prostate AFMS \\
        \cellcolor{acolor}A19: Rectum & \cellcolor{hcolor}H25: Larynx glottis & \cellcolor{pcolor}P01: Gluteus maximus $\ast$ & \cellcolor{lcolor}L19: Hypoxic-ischemic  \\
        \cellcolor{acolor}A20: Portal veins & \cellcolor{hcolor}H26: Larynx & \cellcolor{pcolor}P02: Gluteus medius $\ast$ & \cellcolor{lcolor}L20: Breast tumor \\
        \cellcolor{acolor}A21: Large bowel & \cellcolor{hcolor}H27: Pharyngeal constrictor & \cellcolor{pcolor}P03: Gluteus minimus $\ast$ & \cellcolor{lcolor}L21: Glioma \\
        \cellcolor{hcolor}H01: Brain & \cellcolor{scolor}S01: Clavicle $\ast$ & \cellcolor{pcolor}P04: Bladder & \cellcolor{lcolor}L22: Thyroid nodule \\
        \cellcolor{hcolor}H02: Face & \cellcolor{scolor}S02: Femur $\ast$ & \cellcolor{pcolor}P05: Prostate and uterus & \cellcolor{lcolor}L23: Skin lesion \\
        \cellcolor{hcolor}H03: Airway & \cellcolor{scolor}S03: Hip $\ast$ & \cellcolor{pcolor}P06: Prostate & \cellcolor{lcolor}L24: Polyp \\
        \cellcolor{hcolor}H04: Eye $\ast$ & \cellcolor{scolor}S04: Humerus $\ast$ & \cellcolor{pcolor}P07: Testicle & \cellcolor{pcolor}P10: Prostatic urethra \\
        \cellcolor{hcolor}H05: Crystalline lens $\ast$ & \cellcolor{scolor}S05: Rib (L\&R, 1-12) & \cellcolor{pcolor}P08: Prostate peripheral zone & \\
        \cellcolor{hcolor}H06: Optic nerve $\ast$ & \cellcolor{scolor}S06: Sacrum & \cellcolor{pcolor}P09: Prostate transition zone & \\
        \bottomrule
    \end{tabular}
    \label{sup_tab1}
\end{table}

\textbf{Division of Datasets.} Tab.\ref{fig2} summarizes the datasets used for training and testing our model. Each dataset was divided into 90\% for training and 10\% for testing, with 3D datasets split along the volume dimension. To ensure evaluation reliability, we limited the test set of each dataset to a maximum of 3,000 images for final model assessment. Additionally, we evaluated the model's zero-shot capability using three external datasets: SegThor~\cite{100lambert2020segthor}, TotalSegmentatorMRI~\cite{Wasserthal_2023}, and ISLES~\footnote{https://www.isles-challenge.org/}. Therefore, our training and test sets share the same data distribution, including modality and category.

\begin{center}
\scriptsize
\setlength{\tabcolsep}{1pt} 
\begin{longtable}{p{4cm}|p{6cm}p{2cm}p{1.3cm}} 
\caption{\textbf{Training and Test Datasets.} The following datasets were collected for training and validating IMIS-Net. Processed non-private datasets will be made publicly available for research purposes.} \label{sup_tab2} \\
\hline

\textbf{Dataset} & \textbf{Segmentation target} & \textbf{Modality} & \textbf{Category} \\ \hline
\endfirsthead

\hline
\multicolumn{4}{c}{{\tablename\ \thetable{} : Continued from previous page}} \\ \hline
\textbf{Dataset} & \textbf{Segmentation target} & \textbf{Modality} & \textbf{Category} \\ \hline
\endhead

\hline
\multicolumn{4}{r}{{Continued on next page}} \\ \hline
\endfoot

\hline
\endlastfoot

SegRap2023~\cite{luo2023segrap2023} & A18; H01,04-20,25-26; T01,09 & CT & 45 \\
\rowcolor{gray!20}
AbdomenAtlasMini1.0~\cite{qu2024abdomenatlas} & A02,06,10-13,16-17 & CT & 9 \\
AbdomenCT1K~\cite{ma2021abdomenct} & A11-13,16 & CT & 4 \\
\rowcolor{gray!20}
AMOS2022~\cite{ji2022amos} & A01-02,05-06,10-13,16-17; P04-05; T01 & CT & 15 \\
BTCV~\cite{landman2015miccai} & A01-02,06,10-14,16-17; T01 & CT & 13 \\
\rowcolor{gray!20}
Colorectal\_Liver\_Metastases~\cite{martin2020colorectal} & A12; L02 & CT & 2 \\
Continuous\_Registration\_task1~\cite{marstal2019continuous} & T10 & CT & 1 \\
\rowcolor{gray!20}
COVID-19 CT scans~\cite{paiva2020helping, jun2020covid} & T10; L01 & CT & 1 \\
CTSpine1K\_Full~\cite{deng2021ctspine1k} & S-08-10 & CT & 25 \\
\rowcolor{gray!20}
Finding-lungs-in-cTdata\_3d~\cite{kuijf2019standardized} & T10 & CT & 1 \\
FLARE21~\cite{MedIA-FLARE21} & A11-13,16 & CT & 4 \\
\rowcolor{gray!20}
FLARE22~\cite{FLARE22} & A01-02,05-06,10-13,16-17; T01 & CT & 13 \\
HCC-TACE-Seg~\footnote{https://www.cancerimagingarchive.net/collection/hcc-tace-seg/} & A02,12,20; L02 & CT & 4 \\
\rowcolor{gray!20}
KiTS~\cite{heller2019kits19} & A11; L03 & CT & 2 \\
KiTS2021~\cite{zhao2021coarse} & A11; L03-04 & CT & 3 \\
\rowcolor{gray!20}
KiTS2023~\cite{myronenko2023automated} & A11; L03-04 & CT & 3 \\
Learn2Reg2022\_AbdomenCTCT & A01-02,06,10-14,16-17; T01 & CT & 13 \\
\rowcolor{gray!20}
Learn2Reg2022\_AbdomenMRCT & A11-12,16 & CT & 4 \\
LITS~\cite{bilic2019liver} & A12; L02 & CT & 2 \\
\rowcolor{gray!20}
LUNA16~\cite{setio2017validation} & T10 & CT & 1 \\
MMWHS~\cite{zhuang2018multivariate, zhuang2016multi, luo2022mathcal} & T02-04,08,14 & CT & 7 \\
\rowcolor{gray!20}
MSD\_Liver~\cite{77antonelli2022medical} & A12; L02 & CT & 2 \\
MSD\_Spleen~\cite{77antonelli2022medical} & A16 & CT & 1 \\
\rowcolor{gray!20}
PleThora & T10; L05 & CT & 2 \\
Prostate-AnatomicaLEdge-Cases~\cite{77antonelli2022medical} & A19; S-02; P04,06 & CT & 5 \\
\rowcolor{gray!20}
SLIVER07~\cite{99heimann2009comparison} & A12 & CT & 1 \\
STACOM\_SLAWT~\cite{karim2018algorithms} & T02 & CT & 1 \\
\rowcolor{gray!20}
Totalsegmentator~\cite{Wasserthal_2023} & A01-17; H01-02; S-01-10; P01-04; T01-09 & CT & 104 \\
VESSEL2012~\footnote{https://zenodo.org/records/8055066} & T10 & CT & 1 \\
\rowcolor{gray!20}
Sz\_cxr~\cite{8477564} & T10 & X-Ray & 1 \\
WORD~\cite{luo2022word} & A01,04-06,11-13,15-17,19; S-02; P04; T01 & CT & 16 \\
\rowcolor{gray!20}
WenYi\_CTA\_Data & A02; T02,04 & CTA & 3 \\
CMRxMotions & T03-04 & MR-CMR & 3 \\
\rowcolor{gray!20}
Myops2020~\cite{87luo2022xmetric,87MyoPS-Net} & T03-04; L06-07 & MR & 5 \\
BraTS2013~\cite{menze2014multimodal, info:doi/10.2196/jmir.2930} & L08-11 & MR-FLAIR & 4 \\
\rowcolor{gray!20}
BraTS2015~\cite{menze2014multimodal, info:doi/10.2196/jmir.2930} & L08-11 & MR-FLAIR & 4 \\
BraTS2018~\cite{menze2014multimodal, bakas2017advancing, bakas2019identifying} & L11-13 & MR-FLAIR & 3 \\
\rowcolor{gray!20}
BraTS2019~\cite{menze2014multimodal, bakas2017advancing, bakas2019identifying} & L11-13 & MR-FLAIR & 3 \\
BraTS2020~\cite{menze2014multimodal, bakas2017advancing, bakas2019identifying} & L11-13 & MR-FLAIR & 3 \\
\rowcolor{gray!20}
BraTS2021~\cite{bakas2017advancing, bakas2019identifying, baid2021rsnaasnrmiccai} & L11-13 & MR-FLAIR & 3 \\
BraTS2023\_GLI~\footnote{https://www.synapse.org/Synapse:syn51156910/wiki/621282} & L08-09,11 & MR-FLAIR & 3 \\
\rowcolor{gray!20}
BraTS2023\_MEN & L09-11 & MR-FLAIR & 3 \\
BraTS2023\_MET & L09-11 & MR-FLAIR & 3 \\
\rowcolor{gray!20}
BraTS2023\_PED & L09-11 & MR-FLAIR & 3 \\
BraTS2023\_SSA & L09-11 & MR-FLAIR & 3 \\
\rowcolor{gray!20}
BraTS-TCGA-GBM~\footnote{https://www.cancerimagingarchive.net/analysis-result/brats-tcga-gbm/} & L11-13 & MR-FLAIR & 3 \\
BraTS-TCGA-LGG & L11-13 & MR-FLAIR & 3 \\
\rowcolor{gray!20}
SPPIN2023~\footnote{https://github.com/myrthebuser/SPPIN2023} & L17 & MR-TIGD & 1 \\
ATLAS2023 & A12; L02 & MR-T1W & 2 \\
\rowcolor{gray!20}
CHAOS\_Task\_4~\cite{CHAOS2021,CHAOSdata2019,kavur2019} & A11-12,16 & MR-T1W & 4 \\
Learn2Reg2022\_AbdomenMRCT & A11-12,16 & MR-T1W & 4 \\
\rowcolor{gray!20}
MSD\_Prostate~\cite{77antonelli2022medical} & P08-09 & MR-T2W & 2 \\
Myops2020~\cite{87luo2022xmetric,87MyoPS-Net} & T03-04,06-07 & MR-T2W & 5 \\
\rowcolor{gray!20}
CHAOS\_Task\_4~\cite{CHAOS2021,CHAOSdata2019,kavur2019} & A11-12,16 & MR-T2W & 4 \\
ISBI-MR-Prostate-2013~\footnote{https://www.cancerimagingarchive.net/analysis-result/isbi-mr-prostate-2013/} & P06,08 & MR-T2W & 2 \\
\rowcolor{gray!20}
Prostate\_MRI~\cite{77antonelli2022medical} & P06 & MR-T2W & 1 \\
PROSTATEx-Seg-HiRes~\footnote{https://www.cancerimagingarchive.net/analysis-result/prostatex-seg-hires/} & P06 & MR-T2W & 1 \\
\rowcolor{gray!20}
PROSTATEx-Seg-Zones & P08-10; L18 & MR-T2W & 4 \\
u-RegPro~\footnote{https://muregpro.github.io/} & P06 & MR-T2W & 1 \\
\rowcolor{gray!20}
ADAM2020 & L16 & MR-TOF & 2 \\
ACDC~\cite{8360453} & T03-04 & MR & 3 \\
\rowcolor{gray!20}
AMOS2022~\cite{ji2022amos} & A01-02,05-06,10-13,16-17; P04-05; T01 & MR & 15 \\
EMIDEC~\cite{lalande2022deep} & T03-04; L14-15 & MR & 4 \\
\rowcolor{gray!20}
Heart\_Seg\_MRI~\cite{tobon2015benchmark} & T02 & MR & 1 \\
MMWHS~\cite{zhuang2018multivariate, zhuang2016multi, luo2022mathcal} & T02-04,08,14 & MR & 7 \\
\rowcolor{gray!20}
Mnms2 & T03-04 & MR & 3 \\
MSD\_Heart~\cite{78simpson2019large} & T02 & MR & 1 \\
\rowcolor{gray!20}
PROMISE12~\cite{91litjens2014evaluation} & P06 & MR & 1 \\
CETUS2014~\footnote{https://www.creatis.insa-lyon.fr/Challenge/CETUS/} & T04 & US & 1 \\
\rowcolor{gray!20}
CuRIOUS2022\_tumor & L19 & US & 1 \\
TDSC-ABUS2023~\footnote{https://tdsc-abus2023.grand-challenge.org/} & L20 & US & 1 \\
\rowcolor{gray!20}
u-RegPro & P06 & US & 1 \\
BraimMRI~\cite{braimMRI} & L21 & MR & 1 \\
\rowcolor{gray!20}
Brain-MRI & L21 & MR & 1 \\
UW-Madison & A15,17,21 & MR & 3 \\
\rowcolor{gray!20}
DDTI & L22 & US & 1 \\
drishti\_gs\_cup~\cite{sivaswamy2015comprehensive, sivaswamy2014drishti} & H23 & Fundus & 1 \\
\rowcolor{gray!20}
drishti\_gs\_od~\cite{sivaswamy2015comprehensive, sivaswamy2014drishti} & H24 & Fundus & 1 \\
gamma~\cite{8252743,orlando2020refuge,fu2020age} & H23-24 & Fundus & 2 \\
\rowcolor{gray!20}
ichallenge\_adam\_task2~\cite{} & H24 & Fundus & 1 \\
PAPILA~\cite{kovalyk2022papila} & H23-24 & Fundus & 2 \\
\rowcolor{gray!20}
refuge2~\cite{ORLANDO2020101570, li2020development} & H23-24 & Fundus & 2 \\
rimonedl & H23-24 & Fundus & 2 \\
\rowcolor{gray!20}
finding-lungs-in-cTdata\_2d~\cite{kuijf2019standardized} & T10 & CT & 1 \\
isic2016\_task1~\footnote{https://challenge.isic-archive.com/data/} & L23 & Dermoscopy & 1 \\
\rowcolor{gray!20}
isic2017\_task1 & L23 & Dermoscopy & 1 \\
isic2018\_task1 & L23 & Dermoscopy & 1 \\
\rowcolor{gray!20}
ph2~\cite{93mendoncca2013ph} & L23 & Dermoscopy & 1 \\
cvc\_clinicdb~\cite{bernal2015wm} & L24 & Endoscopy & 1 \\
\rowcolor{gray!20}
endovis15~\cite{bernal2017comparative} & L24 & Endoscopy & 1 \\
hyper-kvasir-segmented-images & L24 & Endoscopy & 1 \\
\rowcolor{gray!20}
kvasir\_seg~\cite{jha2020kvasir} & L24 & Endoscopy & 1 \\
kvasir\_seg\_aliyun & L24 & Endoscopy & 1 \\
\rowcolor{gray!20}
kvasircapsule\_seg~\cite{jha2021nanonet} & L24 & Endoscopy & 1 \\
sun\_seg & L24 & Endoscopy & 1 \\
\rowcolor{gray!20}
SegRap2023~\cite{luo2023segrap2023} & A18; H01,04-20,25-26; T01,09 & CTA & 45 \\
Private1 & A19; S-02; P04 & CT & 4 \\
\rowcolor{gray!20}
Private2  & A01-17; H01-02; S-01-10; P01-04; T01-09 & CT & 104 \\
Private3 & A18; H09-10 & CT & 4 \\
\rowcolor{gray!20}
Private4 & A01-17; H01; S-01-10; P01-04; T01-09 & CT & 104 \\
Private5 & H01 & CT & 1 \\
\rowcolor{gray!20}
Private6 & H11, 13-14 & CT & 10 \\
Private7 & H15-17,25,27 & CT & 6 \\
\rowcolor{gray!20}
Private8 & T01,09-11,18 & CT & 6 \\
Private9 & A11-12,16-17 & CT & 5 \\
\rowcolor{gray!20}
Private10 & A01-17; H01-02; S-01-10; P01-04; T01-09 & CT & 104 \\
Private11 & A19; S-02; P04,06,07 & CT & 6 \\
\rowcolor{gray!20}
Private12 & T13 & CT & 2 \\
Private13 & T02,04,09,12 & CT & 7 \\
\rowcolor{gray!20}
Private14 & A01-19; H01-14; S-01-10; P01-04; T01-09 & CT & 130 \\
\hline
\end{longtable}
\end{center}

\begin{figure}[h]
    \centering
    \includegraphics[width=0.95\linewidth]{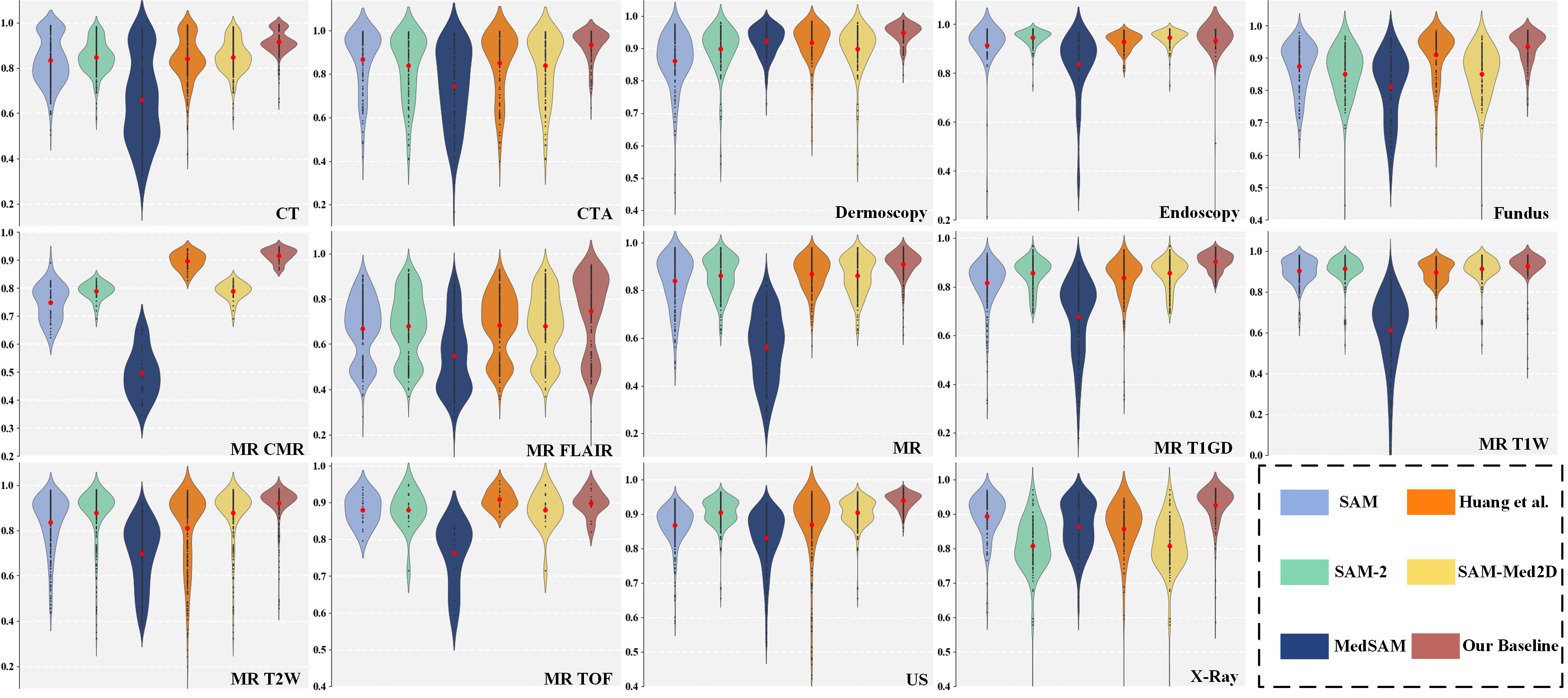}
    \caption{Comparison of segmentation performance of different methods in 14 medical image modalities, where the red points represent the means.}
    \label{sup_fig3}
\end{figure}

\section*{Appendix C: Additional Experimental Analysis}

\textbf{Comparative experiments with other models on different modalities.} Fig.\ref{sup_fig3} shows the segmentation performance across 14 medical image modalities, with the red dots representing the average values. The methods proposed by MedSAM and Huang et al. utilize bounding box prompts, while the other models are evaluated based on better performance achieved from either bounding box prompts or three-click inputs. It can be observed that for medical modalities similar to natural images (e.g., dermoscopy and endoscopy), the performance of SAM and SAM-2 is comparable to the fine-tuned models, which validates the effectiveness of large-scale pretraining data. Additionally, our baseline model performs excellently across 12 modalities, with Dice scores exceeding 90\%, and shows significant stability in modalities such as CTA, dermoscopy, ultrasound, and X-ray. These results suggest that directly using SAM or SAM-2 as a solution for medical image modalities with similar characteristics to natural images is feasible. However, for modalities that significantly differ from natural images, fine-tuning the base model can significantly improve segmentation performance.

\begin{figure}[h]
    \centering
    \includegraphics[width=0.95\linewidth]{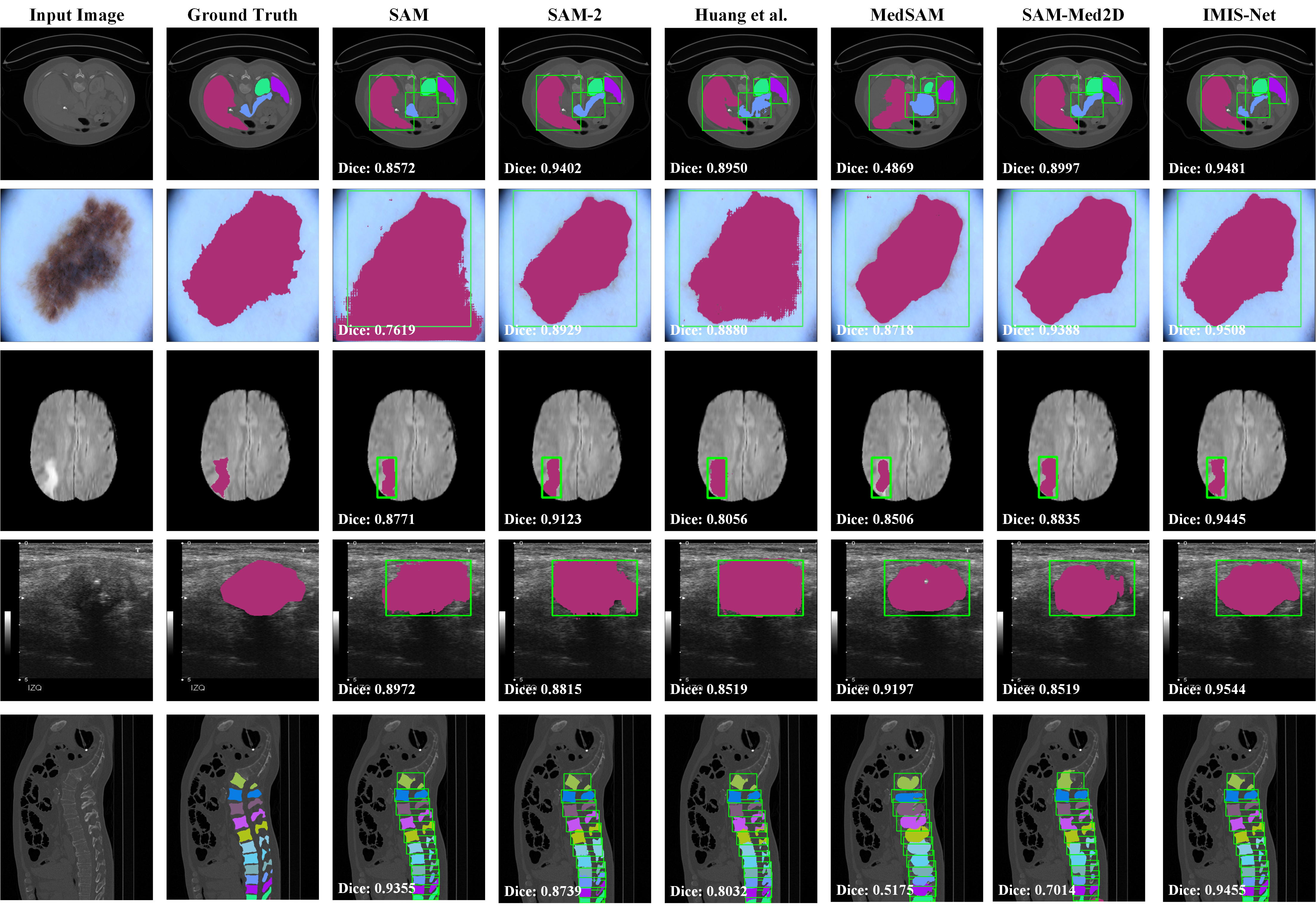}
    \caption{Simulate interactive segmentation results with identical bounding box coordinates for all model inputs.}
    \label{sup_fig4}
\end{figure}

\begin{figure}[h]
    \centering
    \includegraphics[width=0.95\linewidth]{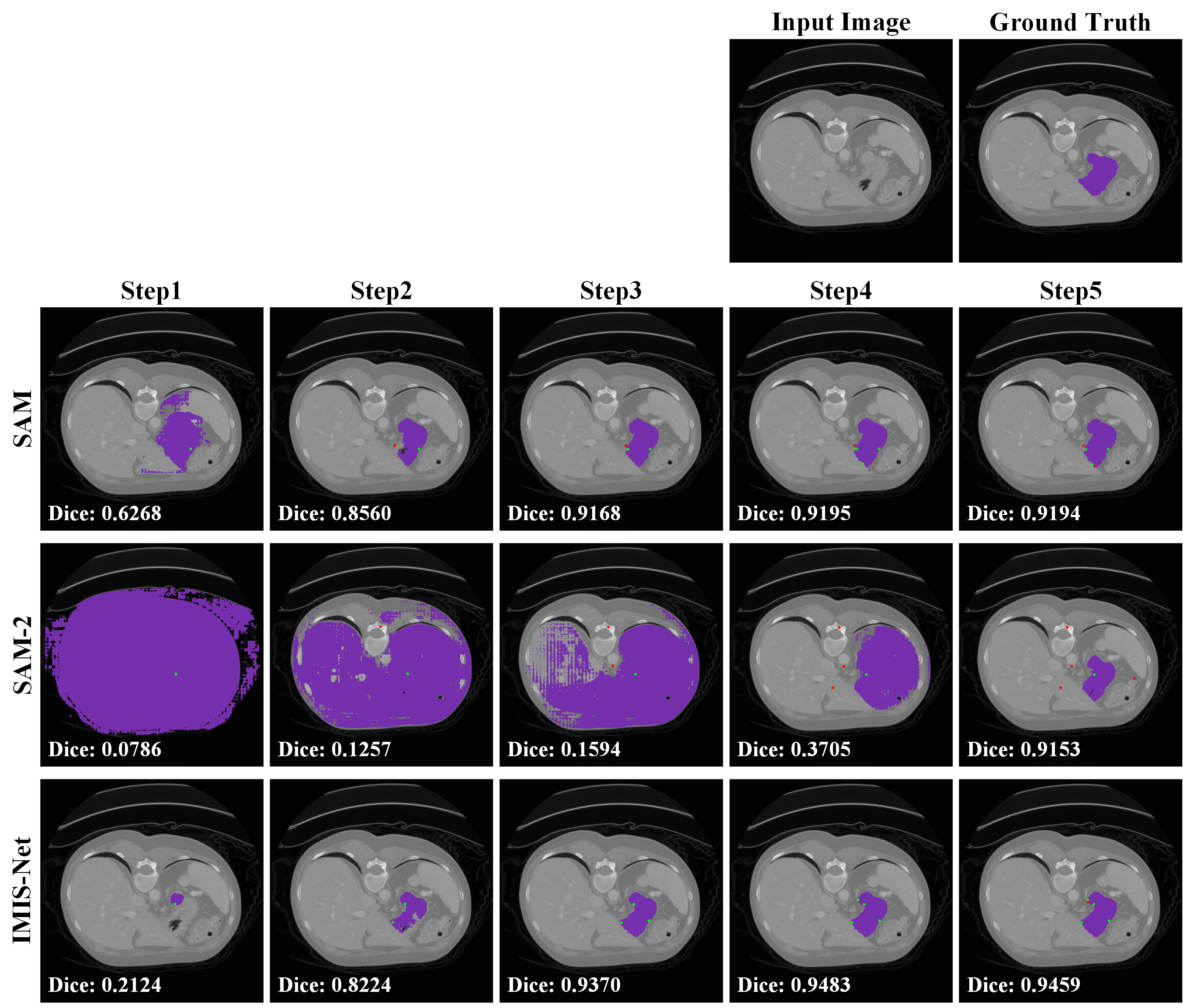}
    \caption{An interactive segmentation example of the stomach in CT images. SAM and SAM-2 typically require more prompts to achieve better results, while IMIS-Net achieves comparable performance with fewer interactions.}
    \label{sup_fig5}
\end{figure}

\begin{figure}[h]
    \centering
    \includegraphics[width=0.95\linewidth]{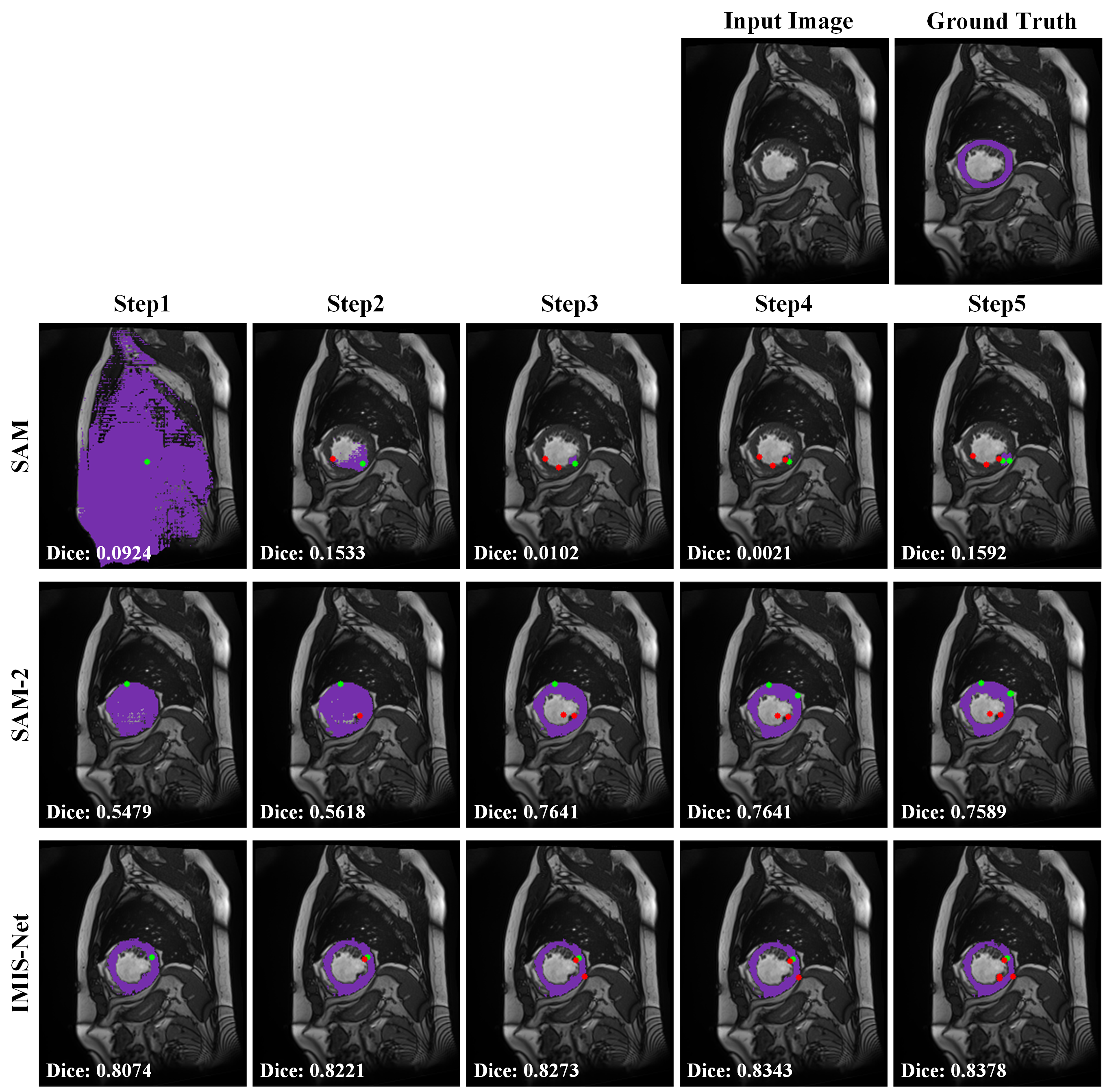}
    \caption{An interactive segmentation example of the cardiac myocardium in MR images. SAM performs poorly when dealing with annular myocardium, while SAM-2 and IMIS-Net are able to obtain predictions of the target area through multiple interactions. Our network consistently outperforms other methods.}
    \label{sup_fig6}
\end{figure}

\textbf{Visualization of interactive segmentation.} We assess interactive segmentation performance using Dice scores. The minimum enclosing bounding box of the ground truth serves as the model's prompt input. Fig.\ref{sup_fig4} presents the prediction results generated by different models based on a single bounding box prompt. Due to the detailed spatial information provided by the bounding box, the Dice scores of various models are mostly above 0.8. In practical applications, this singular interactive approach may not directly meet user needs; hence, the IMIS method supports correcting predictions by providing additional click interactions. Our IMIS-Net still achieves high Dice segmentation. As shown in Fig.\ref{sup_fig5} and Fig.\ref{sup_fig6}, we visualize the results of 5 simulated interactive experiments for SAM, SAM-2, and IMIS-Net.

\renewcommand{\bibsection}{%
  \section*{\refname}
  \small 
}

\clearpage 

\end{document}